\documentclass[11pt,letterpaper]{article}
\usepackage{cogsys}
\usepackage[T1]{fontenc}
\usepackage{times}
\usepackage[pdftex]{graphicx} % use this when importing PDF files
\usepackage{xspace}
\usepackage{booktabs}
\usepackage{graphicx}  %Required
\usepackage{amssymb}
\usepackage{amsmath}

\usepackage{natbib}
\usepackage{comment}

\usepackage{color}
\usepackage{url}
\usepackage{multirow}

\usepackage{array}
\newcounter{rowno}
\cogsysheading{11}{2021}{1--**}{9/2021}{11/2021}

\ShortHeadings{Self-directed Learning of Action Models}{D. \ DANNENHAUER, M. \ FLOYD, N. \ REIFSNYDER, M. \ MOLINEAUX,  D. \ AHA}

\newcolumntype{?}{!{\vrule width 1pt}}

\usepackage{algorithm}
\usepackage[noend]{algpseudocode}

\usepackage{verbatimbox}

\usepackage{todonotes} % for comments

\usepackage{atbegshi}

\DeclareMathOperator*{\argmin}{arg\,min}

\algdef{SE}[DOWHILE]{Do}{doWhile}{\algorithmicdo}[1]{\algorithmicwhile\ #1}%

\begin{document}
%\KER{1}{24}{00}{0}{2004}{S000000000000000}

%\runningheads{DANNENHAUER, FLOYD, REIFSNYDER, MOLINEAUX,  AHA}{Self-directed Learning of Action Models using Exploratory Planning}

%\doublespacing

\title{Self-directed Learning of Action Models using Exploratory Planning}

\author{Dustin Dannenhauer}{dustin.dannenhauer@parallaxresearch.org}
\author{Matthew Molineaux}{matthew.molineaux@parallaxresearch.org}
\address{Parallax Advanced Research, Beavercreek, OH USA}
\author{Michael W. Floyd}{michael.floyd@knexusresearch.com}
\address{Knexus Research Corporation, National Harbor, MD USA}
\author{Noah Reifsnyder}{ndr217@lehigh.edu}
\address{Lehigh University, Bethlehem, PA USA}
\author{David W. Aha}{david.aha@nrl.navy.mil}
\address{Naval Research Laboratory, Washington, D.C. USA}

\vskip 0.2in

\begin{abstract}
Complex, real-world domains may not be fully modeled for an agent, especially if the agent has never operated in the domain before. The agent's ability to effectively plan and act in such a domain is influenced by its knowledge of when it can perform specific actions and the effects of those actions. We describe a novel exploratory planning agent that is capable of learning action preconditions and effects without expert traces or a given goal. The agent's architecture allows it to perform both exploratory actions as well as goal-directed actions, which opens up important considerations for how exploratory planning and goal planning should be controlled, as well as how the agent's behavior should be explained to any teammates it may have. The contributions of this work include a new representation for contexts called Lifted Linked Clauses, a novel exploration action selection approach using these clauses, an exploration planner that uses lifted linked clauses as goals in order to reach new states, and an empirical evaluation in a scenario from an exploration-focused video game demonstrating that lifted linked clauses improve exploration and action model learning against non-planning baseline agents.
\end{abstract}

\section{Introduction}
%\blfootnote{\begin{center}\textbf{Distribution Statement `A' (Approved for Public Release, Distribution Unlimited)}\end{center}}
Sufficiently complex domains, like those that may be encountered in the real world, are rarely fully defined and modeled in advance. Even minor changes in the domain, like weather conditions, can significantly impact how an agent is able to act. For example, the action preconditions for steering a vehicle in snow differ from the preconditions when the weather is clear. It is unrealistic to assume that a knowledge engineer would be able to provide the agent with information related to all possible domain variants. Instead, it would be advantageous for the agent to be able to dynamically learn how novel domains impact its ability to act and update its action model accordingly.

We present a self-directed exploratory planning approach to learning action models. Our planning agent is capable of operating in new domains, even if those domains differ noticeably from previously encountered domains. We do not assume the agent will have access to an expert-provided domain model or plan traces. Instead, we consider agents that perform exploration to obtain a collection of action interactions and uses those interactions for learning relational action models (i.e., preconditions and effects). We define interactions as triples containing the prior state, action, and post state.

A core concept for the exploration planning and exploration action selection is the notion of a Lifted Linked Clause (LLC), which we discuss more formally in Section \ref{sec:llc}. Briefly, a LLC is a representation of a context in a general manner. For example, a LLC context might be that an agent is in front of a closed door. Regardless of the exact state of the environment or exact type of door, the context of facing a doorway is relevant to which actions the agent can successfully carry out. We define LLCs as lifted subsets of the predicates in the state space where predicates share at least one variable across their arguments. LLCs allow us to represent situations in a general way, such as the agent is in a corner, or the agent is in the same location as an object, without having to refer to an exact state. LLCs often correspond to one or more preconditions for an action in the agent's transition model. By tracking which LLCs are active in the current state, and which actions have been taken when an LLC is active, the agent can prioritize reaching new states with active LLCs the agent has not seen before. This enables the agent to reach new states needed to learn a more complete action model, as our results show in Section \ref{sec:evaluation}. It also increases the rate at which the agent explores the domain.  

The vision for the  agents described here is not expected to be purely exploratory, but instead we expect these agents will operate as members of teams where other teammates may request the agents to pursue provided goals. At any time, the agents may be performing a mixture of exploration (i.e., attempting actions in different contexts), learning (i.e., learning action models), and goal pursuit (i.e., exploiting a learned model to achieve teammate-provided goals). 

This work has five primary contributions:

\begin{itemize}
    \item A formalization of the Lifted Linked Clause (LLC).
    
    \item A novel exploratory action selection approach that prioritizes taking actions that have been attempted the least in the currently active LLCs. 
	\item A novel exploratory planning agent architecture that attempts to reach new states, using LLCs as goals.
	\item A formalism describing the algorithms behind the exploratory action selection and exploratory planning using LLCs.
	\item An evaluation in a scenario inspired from the Dungeon Crawl Stone Soup video game, including a random baseline.
\end{itemize}

The rest of the paper is organized as follows. In Section \ref{sec:relatedwork} we discuss related work, followed by a formalism of LLCs in Section \ref{sec:llc}. Section \ref{sec:agent} presents the architecture of our agent including the algorithms for key components, including the process of inductive learning of action models. We describe our evaluation in Section \ref{sec:evaluation}, results in Section \ref{sec:results} and conclude in Section \ref{sec:conclusions}.  

\section{Related Work}
\label{sec:relatedwork}

Since we are concerned with both exploration and learning relational action models for planning, we draw on prior work from both reinforcement learning (RL) and inductive learning. A difference of this work from RL is that our agent is goal-driven. Given that our agent does not know when actions can be applied (preconditions) and the changes made to the state (effects), we draw on ideas from RL to determine what actions to take in order to obtain enough examples for learning action models.

Hester and Stone (\citeyear{hester2017intrinsically}) discuss multiple exploration reward schemes and reward metrics in intrinsically motivated RL systems. They present two exploration-based intrinsic reward schemes that seek to take the most novel action. Briefly, action novelty takes into account the difference of the current state compared to the most recent state in which an action was executed as well as how uncertain the agents predictions for the state reached after executing the action.  

Sequeira, Melo, and Paiva (\citeyear{sequeira2011emotion}) discuss emotions as a structure for intrinsic motivations which are balanced against extrinsic motivations. Numeric intrinisic motivations are based on appraisal dimensions including novelty, motivation, control, and valence. For example, novelty is proportional to the number of times an action has been used in a given situation before. The use of contexts in our agent is an attempt to reach a similar notion of novelty: taking actions that have not been tried in similar situations. Our agent prioritizes first to take actions that have been taken the least among all the currently active contexts (LLCs, see Section \ref{sec:llc}) and when all actions have been tried, the agent then prioritizes planning to reach states containing new contexts.

Inductive learning approaches that use heuristics have been effective \citep{vere1980multilevel,hayes1978interference,watanabe1990effective} but may not apply to all learning problems. FOIL \citep{quinlan1990learning} is a greedy algorithm that, like the heuristic approaches just mentioned, requires both positive and negative examples for training. Wang (\citeyear{Wang1995}) developed a system called OBSERVER that uses expert traces of actions and a simulator for running practice problems. OBSERVER does not need approximate action models; it learns action models from scratch that are as effective as human-expert coded action models assuming expert traces are available. Our learning problem is similar to those of OBSERVER and FOIL, hence our use of an inductive learning approach. There has been recent work on learning action models using genetic algorithms \citep{kuvcera2018louga} with example plans. The main difference in this work is that we focus on obtaining good examples for learning unknown action models, rather than on the learning algorithm itself. A benefit of the approach we describe is relaxing the need for expert traces of high quality plans.

To perform learning we use the Inductive Learning of Answer Set Programs (ILASP) system \citep{ILASP_system}. ILASP is a powerful inductive learning system that combines brave and cautious induction to learn answer set programs.

To deal with incompleteness of action models, Weber and Bryce (\citeyear{Weber1971}) give a planning algorithm that reasons about the action model's incompleteness in order to avoid failure during planning. Incorporating such a technique into our agent for the goal-driven planner would allow the agent to perform planning prior to the learning of a complete action model (which may not ever happen).

Human-agent teaming, and specifically the need for explanation in such teams, has been identified as a potential challenge problem for Goal Reasoning \citep{molineaux2018}. Their work presents a model for how explanations can be exchanged among teammates and used in decision making, and presents several common instantiations of the model. Our work falls under the \textbf{Single Supervisor} instantiation, where an agent has a supervisor that may make requests of it, and it may need to explain its reasoning or behavior.

Existing applications of explanation to Goal Reasoning have focused primarily on \textit{internal explanations} \citep{aamodt93} (i.e., for the benefit of the agent itself) rather than \textit{external explanations} (i.e., for the benefit of an external user). DiscoverHistory \citep{molineaux2012} generates explanations for why observed domain transitions occurred. More specifically, it generates explanations that contain the external actions (i.e., of other agents) and exogenous event that are most likely to have resulted in the changes in the domain. In a human-robot teaming domain, such explanations have been used to allow a Goal Reasoning agent to determine the plans and goals of other agents, and reason over its own goals \citep{gillespie2015}. However, even though these explanations are used by an agent that is a member of a team, they are only used internally and never provided to any teammates.

\section{Lifted Linked Clause (LLC) Formalism}
\label{sec:llc}

The approaches presented here use \textit{LLCs} to guide the exploration process. A LLC $c$ is a first-order relational conjunct $c_1 \land c_2 \land \ldots \land c_n$ that refers to one or more existentially quantified variables and is satisfied by some states $S' \subset S$. We describe a LLC $c$ having $n$ terms as being of size $n$ (e.g. Figure \ref{all-contexts} shows LLCs of size $n=3,4,5$). LLC terms may include the following: finite relations over objects, finite equalities and inequalities between integers and functions over objects, and the infinite ``successor" relation that describes consecutive integers.  An upper bound\footnote{This upper bound assumes all predicates $p \in P$ have a number of arguments equal to the maximum number of arguments $M$, however usually some $p$ have less than $M$ arguments.} $C$ on the number of possible LLCs for a given size $n$ is given by $$C = {M! \times 2 \times |P| \choose n}$$

\vspace{0.5em}

\noindent where $M$ is the maximum number of arguments for a single predicate $p \in P$. The $2$ is because each predicate can be negated. We define the size $n$ of a LLC as the number of relations (conditions) in the clause. 
In a LLC with multiple predicates, each predicate must refer to at least one existentially quantified variable that is also present in another term; this requirement means that no context of size $n$ is semantically equivalent to the conjunction of two or more smaller contexts of size $< n$. For example, in the LLC of Figure \ref{all-contexts}a, \verb|T1| is such an existentially quantified variable. We describe the set of all legal LLCs for a domain $\Sigma$ up to a size $n$ as $clauses(\Sigma, n)$.

The constraint that each predicate must refer to at least one existentially quantified variable is hypothesized to reduce the space of potential LLCs and follows from an assumption that useful contexts will be defined by predicates that are relationally connected in some way, rather than any arbitrary collection of predicates. We leave for future work additional experiments to evaluate different methods for constraining the LLC space.

%Contexts that do not have at least one shared variable between any two conditions are discarded since we are concerned with contexts for at least one related variable.

%Contexts are generated using the domain model, by choosing all contexts for size $n$ from the permutations of predicates argument types. (We don't need to describe the generation process, just refer to all contexts of size $n$.)

We say that an LLC is \textbf{active} if it unifies with the current state $s$ (more than one unification is possible; as long as there is at least one unification, the LLC is considered active). The agent maintains $\mathcal{L}toA$, a count of the number of times each LLC action-pair $\langle c, a \rangle$ has occurred in the Interaction History, where $\mathcal{L}$ refers to the set of LLCs up to size $n$, $A$ refers to the set of actions, and $c \in \mathcal{L}$. This is given by the following function:

\begin{equation*}	
examples(c,a) = \bigg| \{\langle s_i, a_i, s_{i+1}\rangle \in I \; | \; c \models s_i \land a_i = a\}\bigg| 
\end{equation*} 

\vspace{0.5em}

 Given a planning domain, it is possible to generate all LLCs up to a certain size (in our experiments we choose an LLC size of 2). Although the number of LLCs grows exponentially with $n$, even small LLCs of size 2 can be useful, as shown in our evaluation in Section \ref{sec:evaluation}.

\section{Agent Architecture for Self-Directed Exploration}
\label{sec:agent}
\begin{figure}[H]
	\centering
	\includegraphics[scale=.6]{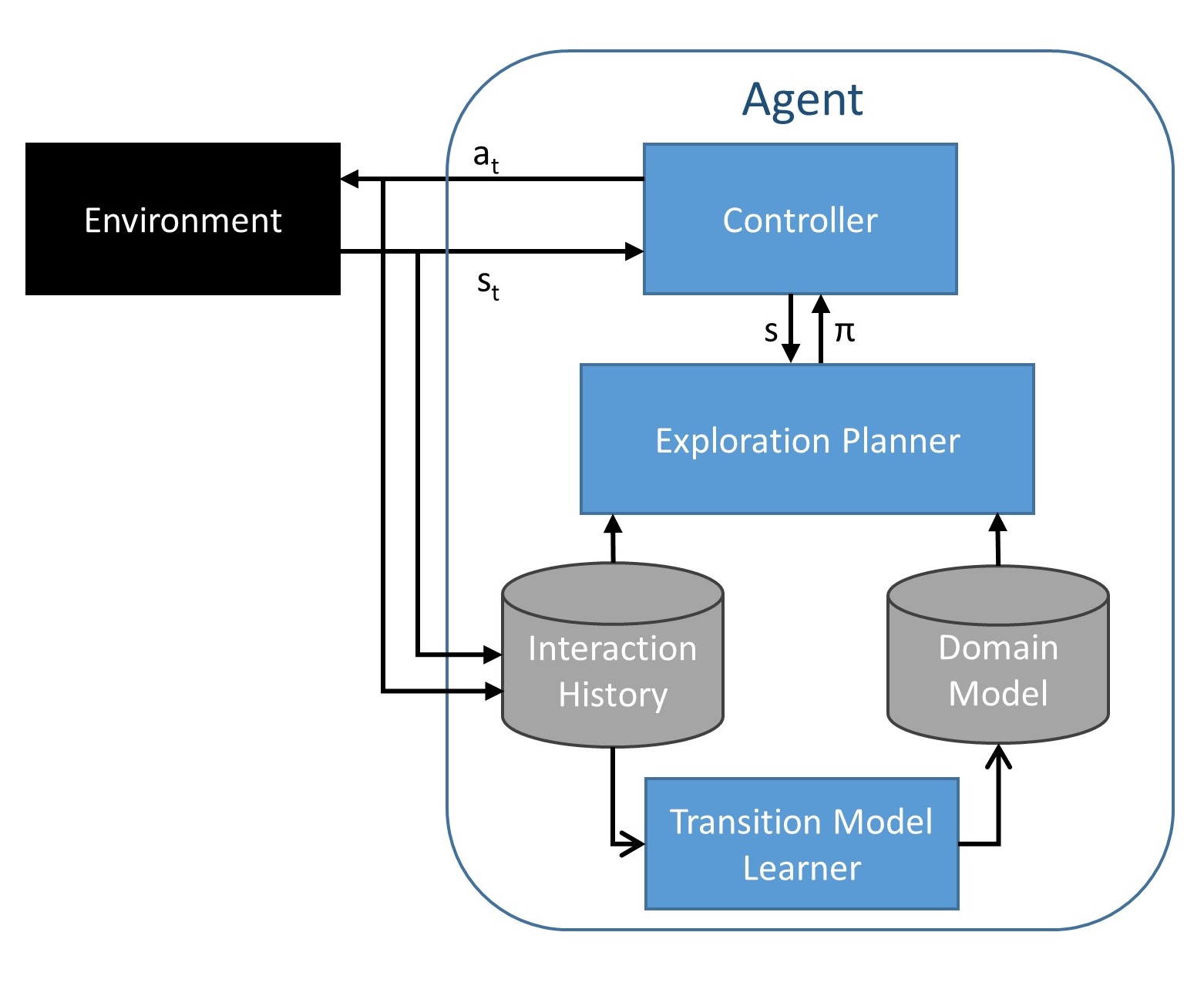}
	\caption{Architecture for our exploratory planning agent}
	\label{fig:agent-design_planning_only}
\end{figure}

We now describe the agent architecture for our self-directed agent. We adopt the planning formalism from \cite{ghallab2004automated} with a state transition system $\Sigma = (S,A,E,\gamma)$ where $S$ is the set of all states, $A$ is the set of actions, $E$ is the set of events, and $\gamma$ is the state-transition function. The architecture for our exploratory planning agent is shown in Figure \ref{fig:agent-design_planning_only}. 

This agent assumes a relational representation of states, observations, and actions are provided to it, but not the transition function. The agent attempts to learn the transition function over time.  

%\note{Matt}{Following paragraphs should describe the purpose of each box in figure 2 in terms of inputs and outputs, followed by anticipated technologies/references for implementation; for example, controller should describe goal selection technologies for use deciding between exploratory/goal-directed planning.}

We separate the functionality of this agent into three submodules. The {\bf exploration planner} is responsible for taking actions to obtain new information with which to update the action model, the {\bf transition model learner}, responsible for updating the agent's model of the world, and the {\bf controller}, responsible for determining when to perform learning (updating the model), planning, or if necessary, take a random action. Each submodule is designed in a  domain-independent fashion, with data represented in a relational fashion. We now describe the current design of each submodule.

% We distinguish exploration planning from goal achievement planning due to the fact that typical goal-based planners seek states, not novelty, and therefore standard heuristics and other optimization's are likely to not be useful for exploration. \dustin{Matt, Floyd: Do you think we should keep the previous sentence? May be a bit strong.}

%We recognize three capabilities for this kind of human-agent team. First, the human needs to identify whether the agent's behavior is correct. Second, if the agent's behavior is correct, the agent needs to clearly explain to the human why its correct. Third, in the case the agent is incorrect, it must be able to explain its behavior to the human in such a way that the human can help the agent fix what is wrong.

%Reinforcement learning is a common paradigm to model an agent that does not know an action model apriori. Q-learning is an example where the agent does not even need to know a transition function. In Q-learning an agent is learning the reward for taking action $a_t$ in state $s_t$, where the reward is the immediate reward plus the expected discounted sum of rewards from this point on assuming optimal actions are always taken. 

\begin{algorithm}[t]
\caption{Controller and Action Selection Process}\label{controller-algorithm}
\begin{algorithmic}[1]
\Procedure{Controller()}{}
\State \textbf{Global:} $s, \mathcal{L}toA$
\While{\textit{running}}
\State $s \gets observe()$ \Comment{Obtain current state}
\State $a \gets ActionSelection()$ \Comment{Choose next action}
\State \For{$c$ in $\mathcal{L}$}
        \If{\textit{active}$(c,s)$} 
            \State $\mathcal{L}toA[c][a].$\textit{increment()} \Comment{Record that this action was taken in this LLC $c$}
        \EndIf
\EndFor 
\State $execute(a)$ \Comment{Execute action $a$}
\EndWhile
\EndProcedure
\Procedure{ActionSelection()}{}
\State \textbf{Global:} $s, \pi, \mathcal{L}toA$
\If{$\pi.length() > 0 $}
\State \Return $\pi.pop()$ \Comment{Retrieve next action of plan}
\EndIf
\State $lta \gets leastTakenActions(s, \mathcal{L}toA)$ \Comment{Retrieve set of actions taken the least\\\hfill in the currently active contexts}
\If{$lta.length() > 0 $}
\State \Return $random.choice(lta)$ \Comment{Return a least taken action}
\Else
\State $\pi \gets explorationPlanner()$ \Comment{Call planning}
\If{$\pi.length() > 0 $}
\State \Return $\pi.pop()$ \Comment{Retrieve next action of plan}
\Else 
\State \Return $randomAction()$ \Comment{Take a random action}
\EndIf
\EndIf
\EndProcedure
\end{algorithmic}
\end{algorithm}

\subsection{Controller} At each time step, the controller receives a state $s_t$ from the domain, then generates a new action $a_t$ to act in the domain. The controller and action selection procedures are given in Algorithm \ref{controller-algorithm}. Lines 3 to 10 describe a common runtime loop where the agent observes the current state (Line 4) and ultimately executes an action (Line 10), and continues indefinitely (Line 3). The exploration approaches (action selection in Algorithm \ref{alg:explore-noplan-algorithm} and planning in Algorithm \ref{planning-algorithm}) must know which actions have been executed in which contexts. This is recorded just before an action is executed by the controller in lines 7-9. The controller calls the action selection function in Line 5 which first checks if there are remaining actions in a plan to be executed, and if so, executes the next action (Lines 13-14). If there are no remaining actions of the plan then the agent may either not be able to generate a plan or finds itself in a new state worth exploring. Line 15 retrieves all actions that have not been taken in currently active contexts (see Algorithm \ref{alg:explore-noplan-algorithm} for the definition of the $leastTakenActions()$ procedure). If the current state is fully explored (in this case the condition on line 17 will be false) there will be no available actions to take, so the agent proceeds to attempt to generate a new plan (line 20) and execute the first action (lines 21-22). Finally, in some situations, the agent must be forced to take a purely random action which occurs on line 24.

The behavior that results for an agent in a new environment with no action model is that it takes random actions until it has learned a partial model. With each successful action taken, the agent finds itself in a different state than previously, and may have actions to take via exploration (lines 15-18). Eventually the agent's model will have learned enough to be able to generate a plan to reach a state with a context that has never been active before. While the plan may be incorrect, in the worst case it will generate new examples for learning, and in the best case it will enable the agent to reach a new state (and new context). From on observer's perspective, by attempting a plan, the agent goes from flailing randomly to directed movement towards a new state.

\subsection{Exploration-based Action Selection using LLCs (part of Controller)}
The exploration-based action selection is shown in Algorithm \ref{alg:explore-noplan-algorithm}. When planning is not possible (such as when the agent may have no model of its action's preconditions or effects) or the agent has found itself in a new situation, this algorithm chooses actions that have not yet been taken in these contexts. Actions are prioritized by the number of contexts they have not been taken in. The algorithm begins by constructing a dictionary tracking the number of active contexts for each action taken (line 3). Lines 4 through 10 iterate over contexts and actions while recording the number of contexts the action has not been taken in. Line 11 creates the variable \textit{leastActions} which keeps track of the actions taken least in currently active contexts. Line 12 introduces a variable tracking the maximum number of contexts an action has not been tried in. This is used to ensure that if there is a tie among actions for number of unknown contexts, then all actions that tie will be returned. Lines 14 through 21 iterate over the actions building the list of actions that have been tried the least. Finally, the actions are return on line 22.

Note that this algorithm is not guaranteed to return a non-empty set of actions. This happens when the agent has fully explored (i.e. executed every action in all of the currently active contexts) the current state. When this occurs the controller performs planning to reach a new state with contexts that have not been fully explored.

\begin{algorithm}[t]
\caption{Exploration-based Action Selection using LLCs}\label{alg:explore-noplan-algorithm}
\begin{algorithmic}[1]
\Procedure{$leastTakenActions()$}{}
\State \textbf{Global:} $\pi, s, \mathcal{L}, \mathcal{L}toA$ 
\State \textit{actionTakenCounts} $\gets Dict()$

\For{$c$ in $\mathcal{L}$}
        \If{\textit{active}$(c,s)$} 
            \For{$a$ in $A$}
                \If{$a \notin $\textit{actionTakenCounts}}
                    \State \textit{actionTakenCounts}$[a] \gets 0$ \Comment{Start tracking this action}
                \EndIf

                \If{$\mathcal{L}toA[c][a] == 0$} \Comment{True if $a$ never taken when $c$ is active}
                    \State \textit{actionTakenCounts}$[a].increment()$ 
                \EndIf
            \EndFor
        
        \EndIf
\EndFor 
\State \textit{leastActions} $\gets []$
\State \textit{max} $\gets 0$ \Comment{Score enables tracking number of contexts\\\hfill that are active and compatible per action}
\For{$a$ in $A$}
    \If{ $a \in actionTakenCounts$ and \textit{actionTakenCounts}$[a] \neq 0$}
        \If{$actionTakenCounts[a] > max$}
            \State $max = actionTakenCounts[a]$
            \State \textit{leastActions} $\gets [a]$ \Comment{Reset \textit{leastActions} and insert this action}
            \State
        \EndIf
        \If{$actionTakenCounts[a] == max$}
        \State \textit{leastActions}$.append(a)$
        \EndIf
    \EndIf
\EndFor

\State \Return $leastActions$

\EndProcedure
\end{algorithmic}
\end{algorithm}

\subsection{Exploration Planner}

Given an initial state $s$, an interaction history $(\langle s_0, a_0, s_1\rangle, \langle s_1, a_1, s_2\rangle, \ldots)$, a domain model of the transition function $\lambda$, and a goal LLC $g$, the exploration planner finds a plan $\pi$ to reach any state that satisfies the goal.

The role of the Exploration Planner is to obtain novel \textit{interactions}. An interaction is defined as a triple $\langle s_i, a_i, s_{i+1}\rangle$ where $s_i \in S$ is the previous state before taking action $a_i \in A$ and $s_{i+1} \in S$ is the state after $a_i$ is executed. As the agent acts in the domain all interactions are recorded and added to the Interaction History $I$.  To determine the next action, the Exploration Planner chooses the action $a$ that minimizes $examples(c,a)$ for all active legal contexts of a size less than or equal to $n$: 

\begin{equation*}
\argmin_{a \in A} \min_{c \in contexts(\Sigma, n)} examples(c,a)
\end{equation*}

\vspace{1em}

We collect complete states from the domain after each action is executed. The Interaction History provides training examples that are later used by the Transition Model Learner to learn a Domain Model. Thus, by guiding the agent's actions, the Exploration Planner can impact which interactions are stored in the Interaction History and used for learning.

\subsection{Transition Model Learner}
We are interested in agents that can learn their action model online while operating in a new domain. For our Transition Model Learner component, we assume that action predicates and argument types are known to the agent a priori (i.e., provided as  expert domain knowledge) as well as a model of the domain that includes objects, types of these objects, and predicates for these objects. Currently we assume the world is fully observable and deterministic; we will relax these constraints in future work.

As we discussed previously, every action the agent takes is recorded into a triple  $\langle s_i, a_i, s_{i+1}\rangle$ and stored in the Interaction History, which can be used later for learning. We have taken an inductive learning approach, inspired by prior work on learning action models. The primary difference in our work is that the agent is not provided with expert traces of action sequences and instead directs its own behavior to acquire examples of executing actions.

\begin{figure}[b]
	\centering
	\begin{verbbox}
	southeast(X, Y) :- agentat(X2, Y2), north(Y2, Y), west(X2, X),
	                   not wall(X, Y), not cdoor(X, Y). 
	\end{verbbox}
	\theverbbox
	\caption{Learned rule from ILASP for the southeast move action in a tile grid world environment.}
	\label{learned_rule_example}
\end{figure}

For the inductive learning algorithm, we use the Inductive Learning of Answer Set Programs (ILASP) system \citep{ILASP_system}  to learn rules that act as preconditions of individual actions.  We assume that if an action did not change the state of the world, then it failed and use that assumption to label interactions as \textit{positive interactions} that succeeded ($s_i \neq s_{i+1}$ after performing $a_i$) or \textit{negative interactions} that failed ($s_i = s_{i+1}$ after performing $a_i$).  Additionally, ILASP is given inductive biases describing the types of predicates that should be considered in the body (in the form of \textbf{\#modeb} declarations), which are translated from the predicates and arguments types from our domain model of our scenarios used in the evaluation.

An example for a precondition rule generated by the ILASP system is shown in Figure \ref{learned_rule_example}. This rule encodes the learned preconditions for the \textit{southeast} move action for a grid-world domain. The various values used in the rule are:
\begin{itemize}
	\item \textit{X}: The X coordinate of the tile the agent is currently located in
	\item \textit{Y}: The Y coordinate of the tile the agent is currently located in
	\item \textit{X2}: The X coordinate of destination tile
	\item \textit{Y2}: The X coordinate of destination tile
\end{itemize}	

In our current setup, we support learning the eight cardinal directions and the only change in the domain that occurs from executing actions is the agent's location or opening / closing a door. 

%This allows us to use Version 2 of ILASP\footnote{ILASP supports inductive learning under various conditions, including a small number of examples (Version 2), an iterative version that scales well with the number of examples (Version 2i), and learning with large numbers of noisy examples (Version 3).} and encode the current state as background knowledge except for the player's location. When we add actions that may change other parts of the state (such as picking up or dropping objects), we will use ILASP's contexts\footnote{This is not related to our use of the term \textit{contexts} in this paper.} feature which allows individual positive or negative examples to be associated with their own background knowledge. This allows for learning rules where some examples are true in some situations and others true in other situations, thus relaxing the requirement of having a single global set of facts (background knowledge) which all rules must be associated with. For full details of ILASP, see \cite{ILASP_system}. 

Before the agent performs planning, the agent makes an external call to the ILASP system to perform learning on all actions that have newly recorded interactions. External calls to ILASP may take up to a few hours to complete, so limiting the number of calls to the ILASP planner should be prioritized. For future work there are a number of ways to reduce ILASP run time including selecting only a subset of interactions for ILASP to learn over. Currently, every interaction, including hundreds of failed interactions because of invalid arguments (i.e. the agent attempts to move to a non-adjacent tile) are given to ILASP. Additionally, ILASP is under active development and future versions may significantly speed up run time. 

\begin{algorithm}[t]
\caption{Exploration-based Planning using LLCs as Goals}\label{planning-algorithm}
\begin{algorithmic}[1]
\Procedure{LLC-Planner}{}
\State \textbf{Global:} $s, A, I, \mathcal{L}toA$
\State $A \gets learnActionModels(I)$ \Comment{Perform learning to ensure up-to-date action models}
\State $G_{done} \gets [$ $]$ \Comment{Store goals already attempted}
\Do
\State $G_{remaining}, G_{done} \gets contextsWithLeastActions(\mathcal{L}toA, G_{done})$ \Comment{Retrieve LLCs}    
\For{$g$ in $G_{remaining}$}
        \State $\pi \gets $\textit{ASP-Planner}$(A,s,g)$ \Comment{Call planner}
            \If{$\pi.length() > 0 $}
            \State \Return $\pi$  \Comment{Return first plan found}
        \EndIf
\EndFor 
\doWhile{$G_{remaining}.length() > 0$} 

\State \Return $[$ $]$

\EndProcedure
\end{algorithmic}
\end{algorithm}

The Transition Model Learner uses an inductive learning procedure to learn the preconditions of each of the agent's actions (i.e., update the Domain Model with the actions' preconditions). An example of the learned preconditions for the \textit{southeast} move action from a tile grid world exploration domain is shown in Figure \ref{learned_rule_example}. The learning component is only concerned with learning action preconditions. Once preconditions are learned, the effects of the actions can be computed by finding the difference between $s_{i+1}$ and $s_i$ and unifying the differences so that the argument variables in the effects clause are consistent with the variables in the preconditions clause.

The intuition behind using contexts for exploration is that the same action needs to be executed in multiple contexts in order to learn the complete precondition rule, and also effects, for that action. Consider the preconditions for \textit{southeast} shown in Figure \ref{learned_rule_example}. In order for the agent to learn the condition \textit{not wall(V0)}, it needs to execute the action in a context where there is a wall southeast of the agent, thereby preventing the action from succeeding. This produces a negative interaction. To learn positive interactions, the agent must execute the action in contexts where there is not a wall in a cell to the southeast. These positive interactions enable learning the other preconditions besides \textit{not wall(V0)}.

%associated with the lowest $k$ among all $\langle c,a \rangle$ provided that $c$ is an active context. Since all $\langle c,a \rangle$ start with $k$ value of 0, the first action is chosen randomly.

\begin{figure}[b]
	\centering
	\begin{verbbox}
		{agent-at(T1), at(T1,X1,Y1), at(T2,X2,Y2)} 
	\end{verbbox}
	\theverbbox \\
	\vspace{1mm}
	\small (a) LLC of size 3 \normalsize
	
	\vspace{2.5mm}
	\begin{verbbox}
		{agent-at(T1), at(T1,X1,Y1), at(T2,X2,Y2), not wall(T2)} 
	\end{verbbox}
	\theverbbox \\
	\vspace{1mm}
	\small (b) LLC of size 4 \normalsize
	
	\vspace{2.5mm}
	\begin{verbbox}
		{agent-at(T1), at(T1,X1,Y1), at(T2,X2,Y2), not wall(T2), X2 = X1-1} 
	\end{verbbox}
	\theverbbox \\
	\vspace{1mm}
	\small (c) LLC of size 5 \normalsize
	
	\caption{Example Lifted Linked Clauses of varying sizes}
	\label{all-contexts}
\end{figure}

%\begin{figure}[b]
%	\centering
%	\begin{verbbox}
%	{agent-at(T1), at(T1,X1,Y1), at(T2,X2,Y2)} 
%	\end{verbbox}
%	\subfloat[LLC of size 3 \label{context3}]{\theverbbox}
%	%\label{context3}
%	
%	\vspace{1mm}
%	\begin{verbbox}
%	{agent-at(T1), at(T1,X1,Y1), at(T2,X2,Y2), not wall(T2)} 
%	\end{verbbox}
%	\subfloat[LLC of size 4\label{context4}]{\theverbbox}
%	
%	\vspace{1mm}
%	\begin{verbbox}
%	{agent-at(T1), at(T1,X1,Y1), at(T2,X2,Y2), not wall(T2), X2 = X1-1} 
%	\end{verbbox}
%	\subfloat[LLC of size 5\label{context5}]{\theverbbox}
%	\caption{Example Lifted Linked Clauses of varying sizes}
%	\label{all-contexts}
%\end{figure}			

%For example, in our custom scenario we have the following five predicates: \textit{at(cell,int,int), wall(cell),agent-at(cell),equal(int, int), sub-equal(int, int, int)} where \textit{sub-equal(x,y,z)} is equivalent to $x = y - z$. The number of argument types $R$ is 2 and the max-args a predicate has is

%The possible contexts of size 3 are: \{center tiles, wall to the east, wall to the north, wall to the south, wall to the west, wall to the northwest, wall to the northeast, wall to the southwest, wall to the southeast\} which is a total of 9 contexts 

%(lifted subsets of states) and the actions that were taken in those contexts. The goals of the exploration planner will be to acquire novel interactions. Over time, the agent may find that some actions have not been taken in some contexts. Thus, the agent 

\section{Evaluation}
\label{sec:evaluation}
\begin{figure}
    \centering
    \begin{minipage}[t][10cm][c]{.5\linewidth}
    \centering
  \includegraphics[scale=0.4]{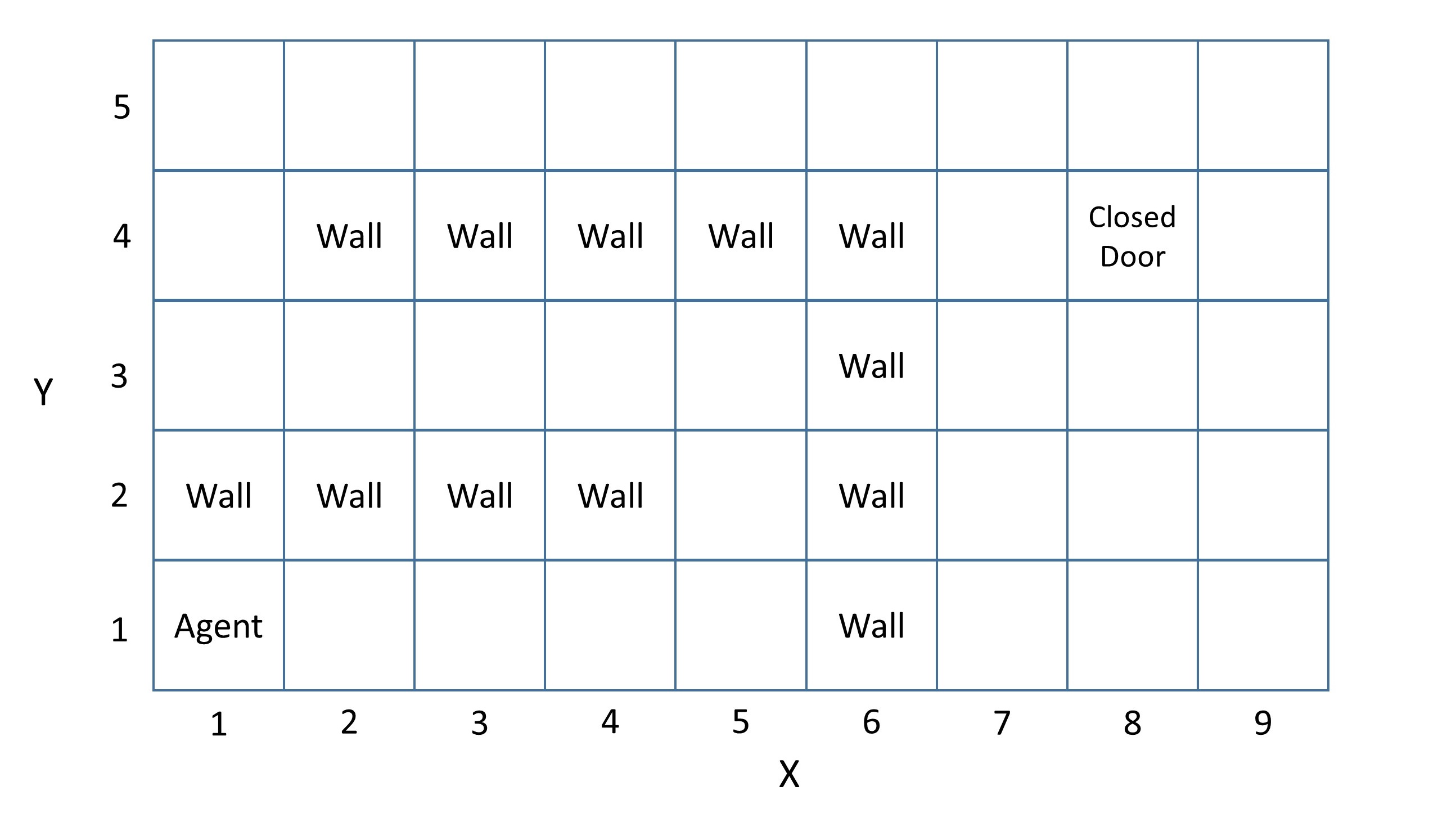}
  
  %\vspace{2.2cm}
  (a) Graphical Representation
  \label{fig:scenario_2_example_graphic}
\end{minipage}% This must go next to `\end{minipage}`
\begin{minipage}[t][10cm][c]{.5\linewidth}
  \centering
  \theverbbox
  
  \vspace{4.5mm}
  (b) PDDL Representation
  \label{fig:scenario_2_example_pdd}
\end{minipage}
    \caption{Scenario 1 Visualization and PDDL Representation}
    \label{fig:scenario1}
\end{figure}

We perform an evaluation of our self-directed planning agent in an ablation study in a scenario inspired from the exploration-based genre of rogue-like video games. In these games humans must figure out when (preconditions) and how (effects) actions change the game state to explore an unknown environment that is partially observable and open. Figure \ref{fig:scenario1} depicts both a graphic visualization and a relational representation of this scenario. We use the Planning Domain Definition Language (PDDL) to represent the state and actions \citep{mcdermott1998pddl}. The PDDL representation of each scenario shown is exactly what is given to our agents as the starting state. Agents are given the actions they can take and typed arguments for these actions without any preconditions or effects. 

The scenario here is static and fully observable. This allows us to label each interaction $\langle s_i, a_i, s_{i+1}\rangle$ as positive if $s_i \neq s_{i+1}$ and negative if $s_i = s_{i+1}$. We leave evaluations in dynamic domains for future work. Briefly, a challenge in a dynamic environment (which is the case for most rogue-like video games) is to identify whether a change in the environment can be attributed to the agent or an external actor. Only in the cases when a change in state occurs and is likely to be caused by the agent should the interaction be considered positive. 

\small
\begin{verbbox}
    (define (problem dcss)
      (:domain dcss)
      (:objects
        x1 x2 x3 x4 x5 x6 x7 x8 x9 - xcoord
        y1 y2 y3 y4 y5 - ycoord)
      (:init
        (agentat x1 y1) (wall x1 y2)
        (north y2 y1)   (wall x2 y2)
        (north y3 y2)   (wall x3 y2)
        (north y4 y3)   (wall x4 y2)
        (north y5 y4)   (wall x2 y4)
        (west x2 x1)    (wall x3 y4)
        (west x3 x2)    (wall x4 y4)
        (west x4 x3)    (wall x5 y4)
        (west x5 x4)    (wall x6 y1)
        (west x6 x5)    (wall x6 y2)
        (west x7 x6)    (wall x6 y3)
        (west x8 x7)    (wall x6 y4)
        (west x9 x8)    (cdoor x8 y4)))
    
\end{verbbox}
\large

Each agent has 24 possible actions consisting of eight movement actions, eight open door actions and eight close door actions. The eight variations of each action type (movement, open door, close door) is because each action is for one of the eight cardinal directions: N, S, E, W, NE, NW, SE, SW. 

\small
\begin{verbbox}
(define (problem dcss)
  (:domain dcss)
  (:objects
    x1 x2 x3 x4 x5 - xcoord
    y1 y2 y3 y4 y5 - ycoord)
  (:init
    (agentat x1 y5) (wall x1 y1)
    (north y2 y1)   (wall x2 y2)
    (north y3 y2)   (wall x3 y3)
    (north y4 y3)   (wall x4 y4)
    (north y5 y4)   (wall x5 y5)
    (west x2 x1)    (wall x3 y2)
    (west x3 x2)    (wall x4 y3)
    (west x4 x3)    (wall x1 y3)
    (west x5 x4)    (wall x2 y4)
    (cdoor x4 y2)   (wall x3 y5)))
    
\end{verbbox}
\large

\subsection{Exploration Progress Metric}
We measure how much of the scenario each agent has explored by counting the number of unique tiles they visited. Visiting more unique tiles means that an agent has explored more of the domain and is a partial indicator of the likelihood of that agent's model being accurate, since without visiting certain tiles, some actions can never be accurately learned (i.e. visiting adjacent tiles to the closed door tile are needed to learn the \textit{open-door} action). Scenario 1 has 33 unique tiles the agent can occupy. These include the closed door tile since if the door is open, the agent can then move into that tile. 

\subsection{Learning Accuracy Metric}
\label{sec:learning_metric}
Since the agents are learning action preconditions from scratch, we develop a metric that compares the learned action models from perfect action models in such a way that the rules do not need to be equivalent regarding the logical representation. Instead, we test each learned precondition for an action against perfect preconditions on a number of test states and compare whether the output of the learned rule (i.e. whether the action can be taken in the test state) is the same as the perfect rule. In this case the output is either that the preconditions hold in which case the action can be taken or the preconditions do not hold and the action will fail to change the state. The perfect action models were hand authored: movement actions contain preconditions to ensure there is a destination cell in that direction and that a wall is not in the destination, closing and opening door actions contain the preconditions there is a destination cell and that there is an open or closed door in the destination, respectively.

\begin{figure}
    \centering
    \includegraphics[scale=0.4]{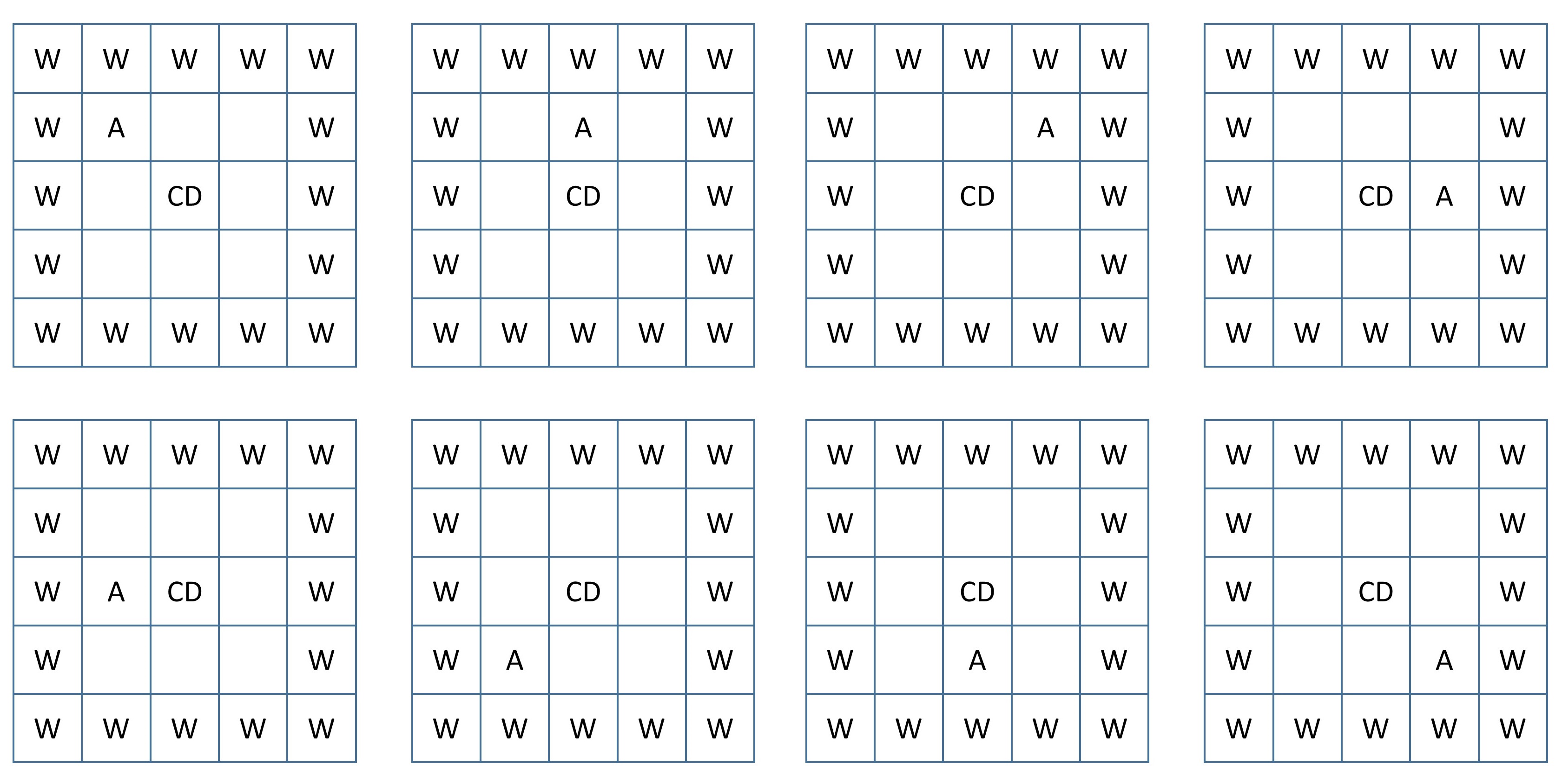}
    \caption{Closed Door Test States Used for Learning Accuracy Metric}
    \label{fig:test_states_closed_door}
\end{figure}

For this evaluation we used 16 expert authored test states. Eight of these states involved a closed door and eight involved an open door. In all test states some movement actions are possible while others are not. The closed door test states are shown in Figure \ref{fig:test_states_closed_door}. \textbf{W} represents a wall, \textbf{A} represents the agent's location, and \textbf{CD} represents a tile with a closed door. Empty tiles are spaces that the agent can move into. An agent may move into a door tile only if the door is open. 

For example, when applying perfect action preconditions on this state the agent is able to successfully execute the following actions: \textit{move south, move east, and close door southeast}. All other actions when applied to this state will fail. Then we apply the learned preconditions for each action. We use the results (i.e. whether or not the perfect/learned rule can be executed in that state) from the test states to calculate precision, recall, and F1 scores between 0 and 100\% for each actions' learned preconditions. The benefit of this metric is that we can measure learning performance irrespective of the rule's representation.

\subsection{Hypotheses}
Our evaluation aims to test whether these agents can learn action preconditions and effects from scratch when situated in a previously unseen domain and if there is a benefit to our exploration planner using LLCs. We hypothesize that:

\begin{description}
	\item[H1] All agents will be able to learn some action preconditions from collected interactions.
	\item[H2] The LLC one-step action selection agent will learn better preconditions for more actions than a baseline agent taking only random actions.
	\item[H3] The LLC one-step action selection agent will explore more of its environment than the random baseline agent.
	\item[H4] The LLC exploratory planning agent will learn better preconditions for more actions than both the one-step LLC action selection agent and the random baseline agent.
	\item[H5] The LLC exploratory planning agent will explore more of its environment than both the one-step LLC action selection agent and the random baseline agent.
\end{description}

\textbf{H1} relates to our claim that the approaches described here can be used to learn action models by collecting interactions, even if the agent is a random baseline. The rational behind \textbf{H2} and \textbf{H3} is that because the one-step LLC agent maintains information about which actions have been taken in which contexts (LLCs) and prioritizes taking actions that have never been taken in particular contexts, it will obtain the needed interactions for learning an action quicker than the random baseline. We expect \textbf{H4} and \textbf{H5} to be true because the agent will eventually choose to plan to reach new states, and this planning will be more efficient than only considering one action ahead.

\begin{figure}[t!]
	\centering
		\includegraphics[scale=0.6]{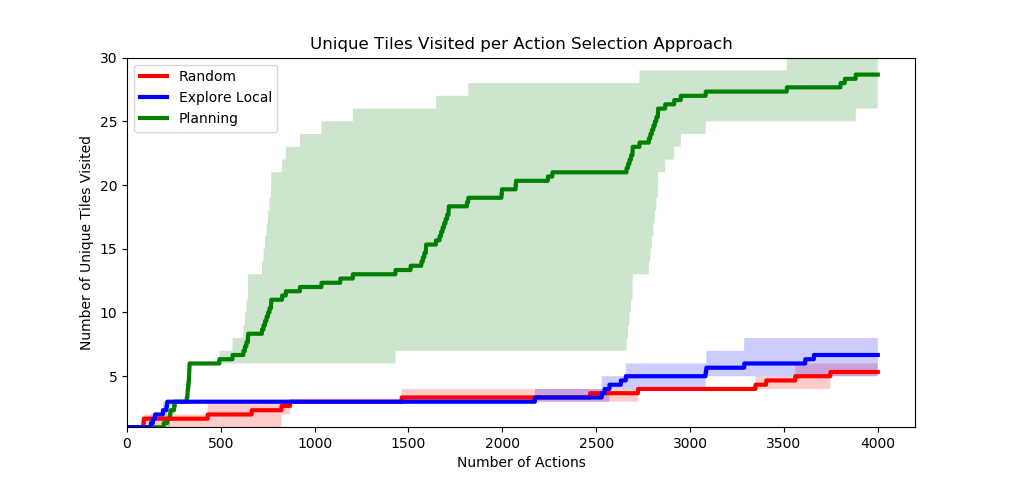}
		\caption{Exploration Progress Per Approach for Scenario 1}
		\label{results_exploration_scenario_1}
\end{figure}

\section{Results}
\label{sec:results}
Our results show the performance of three agents (random, one-step LLC, and planning LLC) in a scenario in Figure 4. The performance of the agents displayed in Figure \ref{results_exploration_scenario_1} are averaged over three runs per agent for 4,000 actions. In each run the agent starts with no preconditions or effects for any of its actions. That is, each run is independent and no knowledge is maintained between runs.

\subsection{Scenario 1 Results}

The graph in Figure \ref{results_exploration_scenario_1} shows the exploration progress of the agent per action executed. Solid lines are composed of the average data point over three runs, where each data point represents the cumulative total number of unique tiles visited by that agent since the beginning of the run. The shaded area shows the maximum and minimum values between all the runs. The larger the shaded area, the more variance there is among the runs.

The random agent explores the slowest initially (within the first 500 actions) and manages to perform as well as the explore local agent (using one-step LLC action selection) until about 2500 actions. Then the explore local agent explores slightly more than the random agent. This shows that our hypothesis \textbf{H3} is true in Scenario 1 although only minimally.

The planning agent in this figure has a slight delay in number of tiles executed but then quickly explores the environment much faster than the other two agents. Even though there is fairly high variance between the runs, even the worst case is still almost entirely better than the other two agents, and this variance reduces by the time the agent has executed approximately 3000 actions (shown by the shaded area reducing in size). This result shows that our hypothesis \textbf{H5} is significantly true in Scenario 1. Finally, there are 33 possible tiles the agents' could visit in Scenario 1, which shows that the planning agent almost explores every location using the same number of actions it takes for the explore local and random agents to visit 7 to 5 tiles respectively.

\begin{table}[t]
    \centering
    \begin{tabular}{|l?c|c|c?c|c|c?c|c|c|}
\hline
& \multicolumn{3}{ c? }{Random} & \multicolumn{3}{ c? }{Explore} & \multicolumn{3}{ c| }{Planning} \\
\hline
& P & R & F1 & P & R & F1 & P & R & F1\\\hline
move\_w & 88 & 100 & 93 & 88 & 100 & 93 & 100 & 100 & 100\\
move\_e & 94 & 100 & 97 & 94 & 100 & 97 & 100 & 100 & 100\\
move\_n & 0 & 0 & 0 & 67 & 29 & 41 & 100 & 100 & 100\\
move\_s & 0 & 0 & 0 & 62 & 67 & 65 & 100 & 100 & 100\\
move\_nw & 33 & 23 & 27 & 67 & 46 & 54 & 65 & 52 & 58\\
move\_ne & 0 & 0 & 0 & 0 & 0 & 0 & 98 & 77 & 86\\
move\_sw & 0 & 0 & 0 & 0 & 0 & 0 & 56 & 50 & 53\\
move\_se & 31 & 33 & 32 & 67 & 46 & 54 & 60 & 48 & 53\\
close\_door\_w & 0 & 0 & 0 & 0 & 0 & 0 & 33 & 33 & 33\\
close\_door\_e & 0 & 0 & 0 & 0 & 0 & 0 & 67 & 67 & 67\\
close\_door\_n & 0 & 0 & 0 & 0 & 0 & 0 & 0 & 0 & 0\\
close\_door\_s & 0 & 0 & 0 & 0 & 0 & 0 & 67 & 67 & 67\\
close\_door\_nw & 0 & 0 & 0 & 0 & 0 & 0 & 0 & 0 & 0\\
close\_door\_ne & 0 & 0 & 0 & 0 & 0 & 0 & 0 & 0 & 0\\
close\_door\_sw & 0 & 0 & 0 & 0 & 0 & 0 & 0 & 0 & 0\\
close\_door\_se & 0 & 0 & 0 & 0 & 0 & 0 & 0 & 0 & 0\\
open\_door\_w & 0 & 0 & 0 & 0 & 0 & 0 & 0 & 0 & 0\\
open\_door\_e & 0 & 0 & 0 & 0 & 0 & 0 & 33 & 33 & 33\\
open\_door\_n & 0 & 0 & 0 & 0 & 0 & 0 & 0 & 0 & 0\\
open\_door\_s & 0 & 0 & 0 & 0 & 0 & 0 & 96 & 100 & 98\\
open\_door\_nw & 0 & 0 & 0 & 0 & 0 & 0 & 0 & 0 & 0\\
open\_door\_ne & 0 & 0 & 0 & 0 & 0 & 0 & 0 & 0 & 0\\
open\_door\_sw & 0 & 0 & 0 & 0 & 0 & 0 & 0 & 0 & 0\\
open\_door\_se & 0 & 0 & 0 & 0 & 0 & 0 & 0 & 0 & 0\\
\hline
\end{tabular}
    \vspace{2mm}
    \caption{Precision, Recall and F1 Score per action model learned by each approach in Scenario 1}
    \label{tab:scenario_1_accuracy}
\end{table}

Turning to Table \ref{tab:scenario_1_accuracy} we see that the random agent was able to learn partial models for four actions, the explore local agent learned partial models for 6 actions, and the planning agent learned models for 13 actions, four of which were fully correct models. The explore local agent was able to learn 2 more models than the random agent, and learned more correct models in all cases where random learned a partial model of the same action, which verifies hypothesis \textbf{H2} in Scenario 1.  Hypothesis \textbf{H4} is almost entirely true for Scenario 1, except when regarding the move southeast action, where the explore agent was able to achieve an F1 score above the planning agent by 1\%. This is due to the explore agent learning a rule with higher precision than the rule learned by the planning agent. Since the planning agent learned significantly more action models than explore, we conclude that this shows hypothesis \textbf{H4} to be true in general.

\section{Conclusions}
\label{sec:conclusions}
We introduce a novel formalism of a situational context which we call a Lifted Linked Clause. LLCs can be used as goals by an agent to learn a more accurate action model by attempting to generate a plan to reach a state with a LLC that has never been seen before. Our evaluation shows that an agent can more efficiently explore a space using LLCs as goals than without such a planning mechanism (i.e. against a non-planning explorer or a random agent). We show that both exploration of the state space and learning accuracy are improved by using such an approach. These agents are able to learn in a single run as soon as they gather a small number (two or more) examples per action for learning. A primary motivation for learning such relational models is that it would enable increased explainability for autonomous planning and acting agents. \\~\\

\begin{acknowledgements} 
\noindent
We thank Michael T. Cox for his comments and support in drafting this article. Additionally, we thank DARPA for supporting this research. The views, opinions and/or findings expressed are those of the authors and should not be interpreted as representing the official views or policies of the Department of Defense or the U.S. Government. 
\end{acknowledgements} 

%\clearpage
\vspace{-0.25in}

{\parindent -10pt\leftskip 10pt\noindent
\bibliographystyle{cogsysapa}
\bibliography{references}

}

% Leave a blank line before the closing brace to ensure the final 
% reference has the proper indentation. 

\end{document}